\documentclass[letterpaper, 10 pt, conference]{ieeeconf}

\IEEEoverridecommandlockouts
\usepackage{url}
\usepackage{graphics}
\usepackage{graphicx}
\usepackage{xcolor}
\usepackage{csquotes}
\usepackage{amsmath}
\usepackage{amssymb}
\usepackage[hidelinks]{hyperref}
\usepackage{booktabs}
\usepackage{balance}
\usepackage{xfp}
\usepackage{xspace}
\usepackage{multirow}
\usepackage{multicol}
\usepackage{cite}

\title{\LARGE \bf
Vernata: Self-Supervised Learning of LiDAR Point Representations
}

\newcommand{\ttgPTMIoU}{56.2}
\newcommand{\ttgPTMAcc}{65.4}

\newcommand{\ttgSonataMIoU}{48.7}
\newcommand{\ttgSonataMAcc}{60.6}
\newcommand{\ttgSonataAcc}{83.6}
\newcommand{\ttgBaselineMIoU}{48.8}
\newcommand{\ttgBaselineMAcc}{63.0}
\newcommand{\ttgBaselineAcc}{83.8}
\newcommand{\ttgSparsityMIoU}{49.6}
\newcommand{\ttgSparsityMAcc}{64.5}

\newcommand{\ttgSinkhornKnoppMIoU}{50.8}
\newcommand{\ttgSinkhornKnoppMAcc}{65.8}

\newcommand{\ttgCMDMIoU}{53.4}
\newcommand{\ttgCMDMAcc}{67.4}

\newcommand{\ttgSparsitySKMIoU}{51.6}
\newcommand{\ttgSparsitySKMAcc}{66.2}

\newcommand{\ttgUltimateMIoU}{54.7}
\newcommand{\ttgUltimateMAcc}{69.0}
\newcommand{\ttgUltimateAcc}{87.0}
\newcommand{\ttgCrdClrMIoU}{50.2}
\newcommand{\ttgCrdClrMAcc}{62.4}

\newcommand{\ttgCrdMIoU}{49.4}
\newcommand{\ttgCrdMAcc}{62.0}

\newcommand{\waymoPTMIoU}{68.3}
\newcommand{\waymoPTMAcc}{77.6}

\newcommand{\waymoSonataMIoU}{43.7}
\newcommand{\waymoSonataMAcc}{58.7}
\newcommand{\waymoSonataAcc}{85.7}
\newcommand{\waymoBaselineMIoU}{49.8}
\newcommand{\waymoBaselineMAcc}{63.3}
\newcommand{\waymoBaselineAcc}{89.1}
\newcommand{\waymoSparsityMIoU}{50.9}
\newcommand{\waymoSparsityMAcc}{64.9}

\newcommand{\waymoSinkhornKnoppMIoU}{51.0}
\newcommand{\waymoSinkhornKnoppMAcc}{65.1}

\newcommand{\waymoCMDMIoU}{56.5}
\newcommand{\waymoCMDMAcc}{69.5}

\newcommand{\waymoSparsitySKMIoU}{51.8}
\newcommand{\waymoSparsitySKMAcc}{65.8}

\newcommand{\waymoUltimateMIoU}{57.1}
\newcommand{\waymoUltimateMAcc}{69.0}
\newcommand{\waymoUltimateAcc}{91.5}
\newcommand{\waymoCrdClrMIoU}{51.2}
\newcommand{\waymoCrdClrMAcc}{64.9}

\newcommand{\waymoCrdMIoU}{50.2}
\newcommand{\waymoCrdMAcc}{63.5}

\newcommand{\customPTMIoU}{19.6}
\newcommand{\customPTMAcc}{24.6}

\newcommand{\customUltimateMIoU}{22.1}
\newcommand{\customUltimateMAcc}{29.3}

\newcommand{\ttgLength}{6,501}
\newcommand{\ttgLengthSSL}{24,722}
\newcommand{\waymoLength}{23,691}
\newcommand{\customLength}{220}

\newcommand{\absdiff}[2]{\fpeval{abs(#1 - #2)}}
\newcommand{\relpercent}[2]{\fpeval{round(100 * abs(#1 - #2) / #2, 1)}\%}

\newcommand{\absWaymoMIoU}{\absdiff{\waymoUltimateMIoU}{\waymoBaselineMIoU}}
\newcommand{\absTtgMIoU}{\absdiff{\ttgUltimateMIoU}{\ttgBaselineMIoU}}
\newcommand{\relpercWaymoMIoU}{\relpercent{\waymoUltimateMIoU}{\waymoBaselineMIoU}}
\newcommand{\relpercTtgMIoU}{\relpercent{\ttgUltimateMIoU}{\ttgBaselineMIoU}}

\newcommand{\printval}[1]{#1\xspace}

\newcommand{\rangewaymoConcertoMIoU}{56.9}
\newcommand{\rangewaymoConcertoMAcc}{69.7}
\newcommand{\rangewaymoConcertoDistMIoUAboveTwenty}{50.1}
\newcommand{\rangewaymoConcertoDistAccAboveTwenty}{63.5}
\newcommand{\rangewaymoMineMIoU}{56.5}
\newcommand{\rangewaymoMineMAcc}{69.5}
\newcommand{\rangewaymoMineDistMIoUAboveTwenty}{50.7}
\newcommand{\rangewaymoMineDistAccAboveTwenty}{64.4}

\newcommand{\rangettgConcertoMIoU}{53.4}
\newcommand{\rangettgConcertoMAcc}{67.2}
\newcommand{\rangettgConcertoDistMIoUAboveTwenty}{48.0}
\newcommand{\rangettgConcertoDistAccAboveTwenty}{55.4}
\newcommand{\rangettgMineMIoU}{53.4}
\newcommand{\rangettgMineMAcc}{67.4}
\newcommand{\rangettgMineDistMIoUAboveTwenty}{48.2}
\newcommand{\rangettgMineDistAccAboveTwenty}{55.2}

\author{Oliver Lemke$^{1, 2}$, Alexander Liniger$^{1}$, Abel Gawel$^{1}$, Marco Hutter$^{1, 2}$%
\thanks{$^{1}$Robotics and AI Institute}%
\thanks{$^{2}$ETH Zurich}%
\thanks{Corresponding author: Oliver Lemke {\tt\small olemke@ethz.ch}}%
}

\begin{document}
\maketitle
\thispagestyle{empty}
\pagestyle{empty}

\begin{abstract}

LiDAR serves as a primary sensing modality for robots operating in outdoor environments.
However, the performance of deep learning models in this domain is severely limited by the scarcity of labeled data, a direct result of the high cost of 3D annotation.
Self-supervised learning addresses this scarcity by learning general-purpose features from unlabeled data.
In this work, we present a multi-modal, multi-teacher distillation framework for self-supervised learning on outdoor LiDAR point clouds.
Building upon the Sonata architecture, we introduce Vernata, consisting of three extensions:
sparse view augmentation to improve robustness against varying point densities, a memory bank mechanism to stabilize resource-constrained training, and cross-modal distillation utilizing dense, high-resolution 2D image features to enable fine-grained semantic guidance.
We evaluate our method on the GrandTour, TartanGround, and Waymo datasets, as well as data collected from our own robotic platforms.
Our experiments demonstrate a significant performance improvement over Sonata baselines, yielding mIoU scores of \printval\ttgUltimateMIoU on TartanGround (+\printval\absTtgMIoU points, +\printval\relpercTtgMIoU) and \printval\waymoUltimateMIoU on Waymo (+\printval\absWaymoMIoU points, +\printval\relpercWaymoMIoU).
Finally, we show that the self-supervised approach maintains strong performance even in reduced-modality settings (lacking color or normals), achieving competitive mIoU scores of \printval\ttgCrdMIoU and \printval\waymoCrdMIoU on the respective datasets.
Our implementation is publicly available at \url{https://github.com/rai-opensource/vernata}.

\end{abstract}

\section{INTRODUCTION}
\label{sec:introduction}

Autonomous robots are increasingly transitioning from research demonstrations in controlled environments to operating in the unstructured real world~\cite{bouman2020autonomous, favaro2023buildingcrediblecasesafety}.
This progress is underpinned by advancements in scene understanding, which equip agents with the semantic context required to perform complex, intelligent actions in unstructured environments~\cite{gu2024conceptgraphs, huang2022visual, liu2025aligning, ma2024survey, lemke2024spot, behrens2025lost}.
LiDAR sensors are central to this advanced perception stack.
Due to their ability to construct metric 3D point clouds~\cite{shan2018topographic}, they have emerged as primary sensors in robotics applications~\cite{zhang2014loam, wisth2022VILENS, sun2020scalability, frey2026grandtour}, frequently fused with camera image data to create a comprehensive semantic-geometric understanding of the robot's surroundings~\cite{vora2020pointpainting, zhuang2021PerceptionAware}.

Driven by fundamental scaling laws~\cite{kaplan2020scaling, bahri2024explaining, hoffmann2022training, baniodeh2025scaling}, the machine learning landscape has shifted towards massive data and compute, a transition that has already revolutionized 2D image understanding~\cite{oquab2023dinov2, simeoni2025dinov3, carion2025sam}.
Yet, the 3D domain has not fully benefited from this trend, as traditional supervised learning on point clouds is heavily constrained by the prohibitive cost of manual data annotation~\cite{behley2019semantickitti, dai2017scannet}.
Self-supervised learning (SSL) offers a compelling alternative to this bottleneck, enabling the learning of general-purpose features directly from raw data~\cite{oquab2023dinov2, simeoni2025dinov3, chen2020simple} that can be efficiently adapted to downstream tasks~\cite{chen2024towards, vorontsov2024foundation, cong2022satmae}.

Historically, however, SSL on point clouds has suffered from performance degradation due to representation collapse~\cite{hou2021exploring, wu2023masked, wu2025sonata}.
The recent Sonata framework~\cite{wu2025sonata} addresses this \enquote{geometric shortcut}, producing high-quality features from point representations.
Motivated by this breakthrough, this work adapts the Sonata approach to the specific domain of outdoor LiDAR point clouds, aiming to enhance performance and robustness in this context.

\begin{figure}
    \centering
    \includegraphics[width=0.98\linewidth]{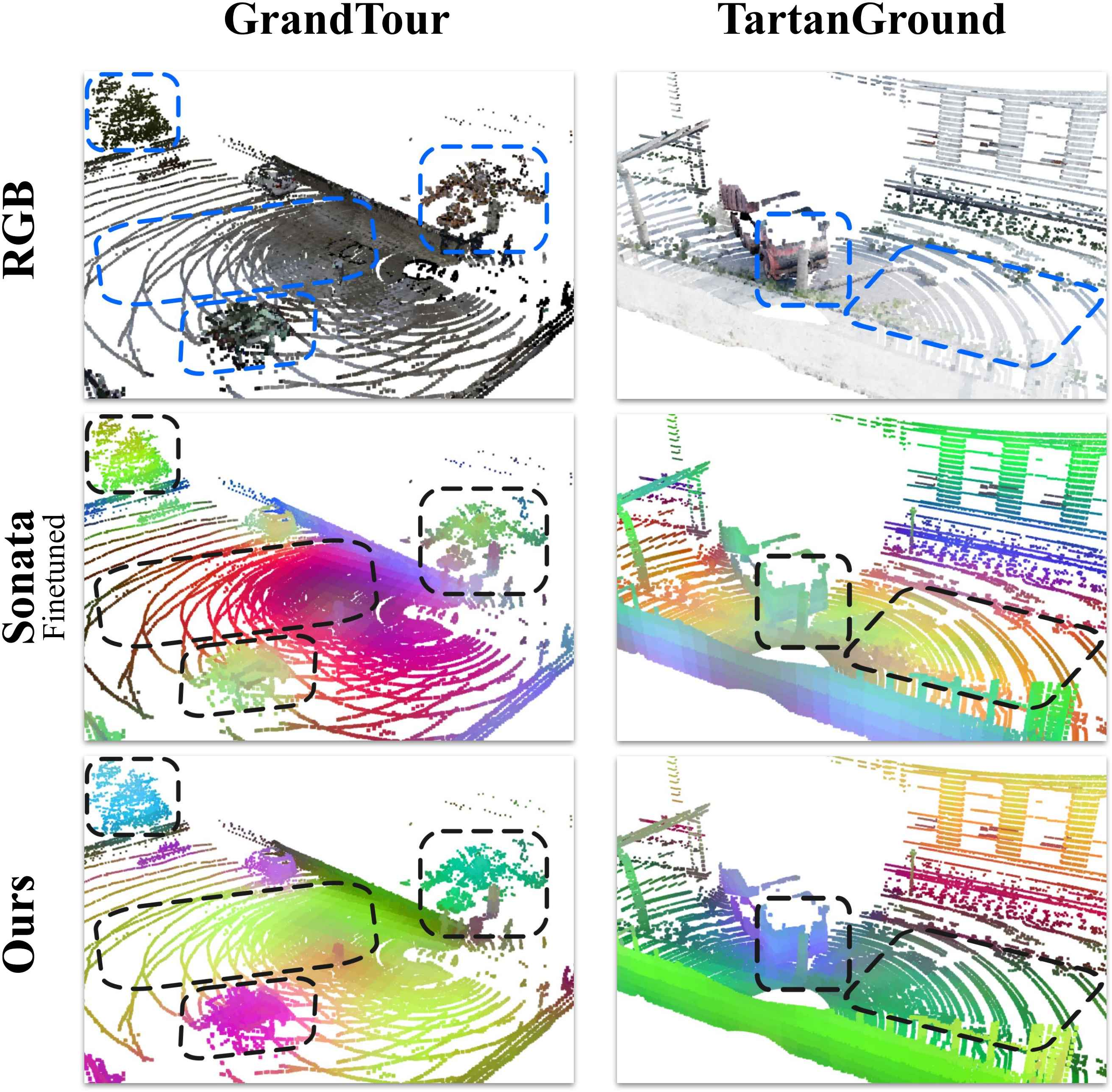}
    \caption{\textbf{Visualization of PCA Features in LiDAR Scenes.}
    We visualize the principal components of point representations produced by Sonata~\cite{wu2025sonata} finetuned on LiDAR datasets, as well as our approach.
    Sonata struggles with separation between distinct semantic classes, as well as consistency across the inherent density variations of LiDAR point clouds.
    Our method integrates cross-modal distillation and a sparse-to-dense objective to overcome these limitations,
    demonstrating sharper semantic distinction, better consistency within object classes, and more robust coherence across planar surfaces like roads, irrespective of the distance to the sensor.
    }
    \label{fig:teaser}
\end{figure}

In this paper, we present Vernata, an extension of the Sonata~\cite{wu2025sonata} framework tailored to the challenges of LiDAR point clouds.
Specifically, we contribute:
\begin{enumerate}
    \item A sparse view augmentation strategy to improve inference robustness under varying density conditions,
    \item The adaptation of memory banks to 3D point cloud representation learning, stabilizing Sinkhorn-Knopp normalization under limited compute constraints, and
    \item The integration of a cross-modal distillation objective using high-resolution image features to enable fine-grained semantic guidance in the LiDAR domain.
\end{enumerate}
We validate our approach via linear probing for semantic segmentation across multiple datasets, including GrandTour~\cite{frey2026grandtour}, TartanGround~\cite{patel2025tartanground}, Waymo~\cite{sun2020scalability}, and a custom real-world collection.
To enable evaluation on TartanGround, we define a custom protocol by selecting scenes with reliable semantic labels, creating a class assignment, and setting up a train/validation split.
We release this benchmark and evaluation protocol to the public.
Our results demonstrate significant improvements over Sonata baselines, achieving mIoU scores of \printval\ttgUltimateMIoU on TartanGround (+\printval\absTtgMIoU points, +\printval\relpercTtgMIoU) and \printval\waymoUltimateMIoU on Waymo (+\printval\absWaymoMIoU points, +\printval\relpercWaymoMIoU).
Finally, we show that the SSL approach maintains high performance even in reduced-modality settings, such as when lacking color or normals, achieving respective mIoU scores of \printval\ttgCrdMIoU and \printval\waymoCrdMIoU.

\section{RELATED WORK}
\label{sec:related_work}

\begin{figure*}[t]
    \centering
    \vspace{0.5em}
    \includegraphics[width=0.75\linewidth]{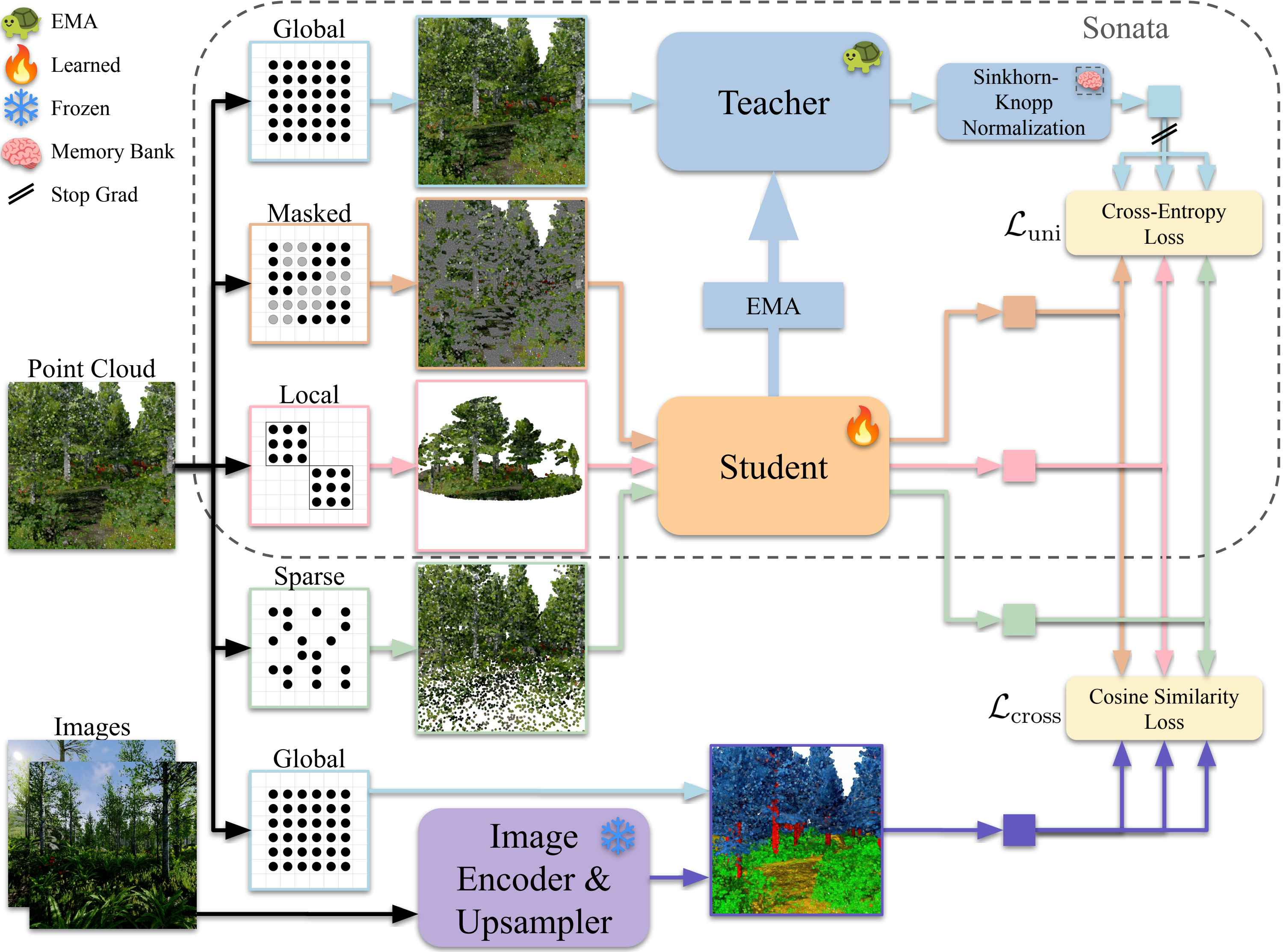}
    \caption{\textbf{Overview of the Framework.}
    We present a multi-modal multi-teacher distillation architecture for self-supervised point cloud learning.
    Building upon the Sonata~\cite{wu2025sonata} framework, highlighted in \textcolor{gray}{gray},
    our model is anchored by two teachers processing \textbf{\textcolor[RGB]{177, 215, 227}{global}} views of the scene: A 3D teacher (\textit{top}), employing a PTv3~\cite{wu2024point} encoder and updated via EMA, and a frozen 2D teacher (\textit{bottom}), producing DINOv2-S~\cite{oquab2023dinov2} \textbf{\textcolor[RGB]{100, 89, 193}{features}}, upsampled with LoftUp~\cite{huang2025loftup}, and backprojected into the point cloud.
    In contrast, the student receives three variations of the global view:
    \textbf{\textcolor[RGB]{235, 181, 145}{masked}} views, \textbf{\textcolor[RGB]{255, 182, 193}{local}} crops, and \textbf{\textcolor[RGB]{186, 215, 185}{sparse}} subsampled views.
    The student embeddings are aligned with the 3D teacher via cross-entropy loss and with the 2D teacher via a cosine similarity-based loss.
    \vspace{-1em}
    }
    \label{fig:method:architecture_main}
\end{figure*}

\textbf{Self-Supervised Learning.}
Motivated by findings that model performance scales effectively with data and compute~\cite{kaplan2020scaling, bahri2024explaining, hoffmann2022training, baniodeh2025scaling}, the field has increasingly adopted self-supervised learning.
Learning representations directly from raw data, SSL allows models to utilize massive unlabeled datasets for scaling~\cite{gui2024survey}.
Early approaches largely relied on contrastive objectives, such as SimCLR~\cite{chen2020simple} and MoCo~\cite{he2020momentum}, which enforce feature invariance across augmented views of the same sample, while pushing apart representations from different samples.
Departing from contrastive instance discrimination, SwAV~\cite{caron2020unsupervised} introduced an online clustering mechanism that enforces consistency by comparing prototype assignments between views.
Subsequently, Masked Image Modeling (MIM), exemplified by MAE~\cite{he2022masked}, demonstrated that reconstructing masked portions of an input encourages deep structural understanding.
More recently, self-distillation methods like DINO~\cite{caron2021emerging} and iBOT~\cite{zhou2021ibot} have combined these insights using a student-teacher framework.
Here, a student network predicts the output of a momentum-updated teacher from local and masked views, a strategy that has successfully scaled to foundation models like DINOv2~\cite{oquab2023dinov2} and DINOv3~\cite{simeoni2025dinov3}.
Our work employs a DINOv2-style architecture while adapting the SwAV~\cite{caron2020unsupervised} memory bank design for point clouds, showing its efficacy in stabilizing limited compute SSL training in the 3D domain.

\textbf{Self-Supervised Learning on Point Clouds.}
Given the high cost of 3D annotation~\cite{behley2019semantickitti, dai2017scannet}, self-supervised learning represents a particularly promising path towards scalable scene understanding in 3D. Early works in this domain mirror the development of the wider SSL field, first focusing on contrastive learning~\cite{xie2020pointcontrast} followed by masked modeling~\cite{wu2023masked}.
Recently, however, Sonata~\cite{wu2025sonata} identified a critical failure mode in these prior methods, coined the \enquote{geometric shortcut}: models regress trivial geometric cues like point height rather than capturing deep semantic meaning.
To address this, Sonata explicitly focuses on obscuring fine-grained spatial information to prevent the model from relying on trivial geometry.
Most notably, the authors remove the decoder network to eliminate high-resolution skip connections that inherently leak geometric details, instead opting for parameter-free up-casting.
In tandem with these architectural changes, Sonata adopts a DINOv2-style~\cite{oquab2023dinov2} self-distillation framework, utilizing a student-teacher setup to enforce local-to-global and masked-to-global consistency.
%
%
Motivated by the success of this paradigm and the recent release of large-scale unlabeled datasets like GrandTour~\cite{frey2026grandtour}, we aim to extend this approach to the outdoor LiDAR domain.
However, this transition is not trivial.
Unlike indoor captures, LiDAR scans exhibit strong range-dependent density variations, overall lower spatial resolution, and often lack auxiliary modalities such as color or reliable surface normals.
To bridge this domain gap, we introduce a sparse-to-dense matching objective to mitigate extreme density variations, alongside high-resolution cross-modal distillation from 2D Vision Foundation Models (VFMs) to encourage semantically consistent representations. 
We also investigate the effect of auxiliary modalities on model performance.

\textbf{Cross-Modal Distillation.}
Integrating 2D semantic information into 3D backbones is a well-established strategy to boost performance~\cite{vora2020pointpainting, bai2022transfusion}.
Beyond explicit fusion~\cite{dai20183dmv, hu2021bidirectional, zhang2023lidar, bai2022transfusion}, cross-modal distillation aims to transfer representations from VFMs to 3D networks without requiring paired images at inference time.
Early methods like PPKT~\cite{liu2021learning} used pixel-to-point contrastive learning, while SLidR~\cite{sautier2022image} improved alignment using superpixels.
Most recently, ScaLR~\cite{puy2024three} demonstrated that simple cosine similarity-based objectives are sufficient for effective scaling when supported by three key pillars: high-capacity 2D teachers, capable 3D students, and diverse training data, a conclusion subsequently corroborated by DITR~\cite{zeid2025dino}.
Our approach integrates cross-modal guidance into the Sonata framework.
Recently, Concerto~\cite{zhang2026concerto} similarly augmented Sonata with 2D feature distillation, matching 3D points to low-resolution DINOv2~\cite{oquab2023dinov2} patches via spatial averaging in the indoor setting.
In contrast, we explore an alternative tailored for the LiDAR domain by investigating point-wise distillation with high-resolution features to preserve fine-grained semantics, especially at range.
Furthermore, rather than bilinearly interpolating the raw low-resolution patches as in ScaLR~\cite{puy2024three}, we first upsample the features with a learned module~\cite{huang2025loftup}.
Ultimately, these works mutually reinforce the value of leveraging 2D VFMs to enhance 3D representation learning.


\section{METHOD}
\label{sec:method}

We present Vernata, a multi-modal, multi-teacher distillation framework designed to learn robust point representations from LiDAR data.
Building upon Sonata~\cite{wu2025sonata},
our approach introduces three extensions to overcome inherent density variations and limited semantic cues in outdoor LiDAR, as well as the practical bottleneck of resource-constrained training.
An overview is shown in Fig.~\ref{fig:method:architecture_main}.

\subsection{Preliminaries: The Sonata Framework}
\label{sec:method:prelim}

Architecturally, Sonata~\cite{wu2025sonata} adapts a DINOv2-style self-distillation framework~\cite{oquab2023dinov2} to the 3D domain.
It employs a dual-network structure consisting of a student and a teacher backbone, where the teacher's parameters are updated via an exponential moving average (EMA) of the student's.

\textbf{View Generation and Objective.}
The training process generates distinct inputs for each branch:
The teacher receives a set of \enquote{global} views $\mathcal{V}_T = \{t_1, t_2\}$, which are large crops covering 40\%-100\% of the scene.
The student receives a set of partial views $\mathcal{V}_S = \mathcal{V}_{\text{mask}} \cup \mathcal{V}_{\text{loc}}$, consisting of \enquote{masked} global views and \enquote{local} crops (5\%-40\% coverage).
To align representations, the student minimizes the cross-entropy between its predicted prototype probabilities $P^{(s)}$ and the teacher's target assignments $Q^{(t)}$ for spatially aligned points.
The targets $Q^{(t)}$ are computed by mapping the teacher's features to learnable prototypes and normalizing the resulting scores via the Sinkhorn-Knopp (SK) algorithm~\cite{caron2020unsupervised, cuturi2013sinkhorn}.
The overall uni-modal objective is defined as:
\begin{equation}
    \label{eq:loss_uni}
    \mathcal{L}_{\text{uni}} = \sum_{s \in \mathcal{V}_S} \lambda_s \sum_{t \in \mathcal{A}(s)} \mathcal{L}_{\text{CE}}(Q^{(t)}, P^{(s)}).
\end{equation}
Here, $\mathcal{A}(s)$ defines the set of teacher views $t$ used as targets for a given student view $s$, and $\lambda_s$ the associated loss weights.
Fig.~\ref{fig:method:architecture_main} highlights the Sonata framework in gray.

\textbf{Geometric Shortcut.}
Sonata not only adapts the DINOv2 architecture to 3D, but also designs its approach specifically to address the \enquote{geometric shortcut}.
The authors identify this as a failure mode unique to 3D SSL, where models collapse to trivial low-level geometry, such as point height or surface normals, rather than learning semantic structures.
To prevent this, Sonata actively obscures fine-grained geometric details by eliminating the standard hierarchical decoder in favor of feature up-casting.
In addition, it employs targeted data augmentations, as well as progressive parameter schedules that gradually increase task difficulty and stabilize training.

\subsection{View Generation with Sparse Augmentation}
\label{sec:method:view_generation}

\begin{figure}
    \centering
    \vspace{0.5em}
    \includegraphics[width=\linewidth]{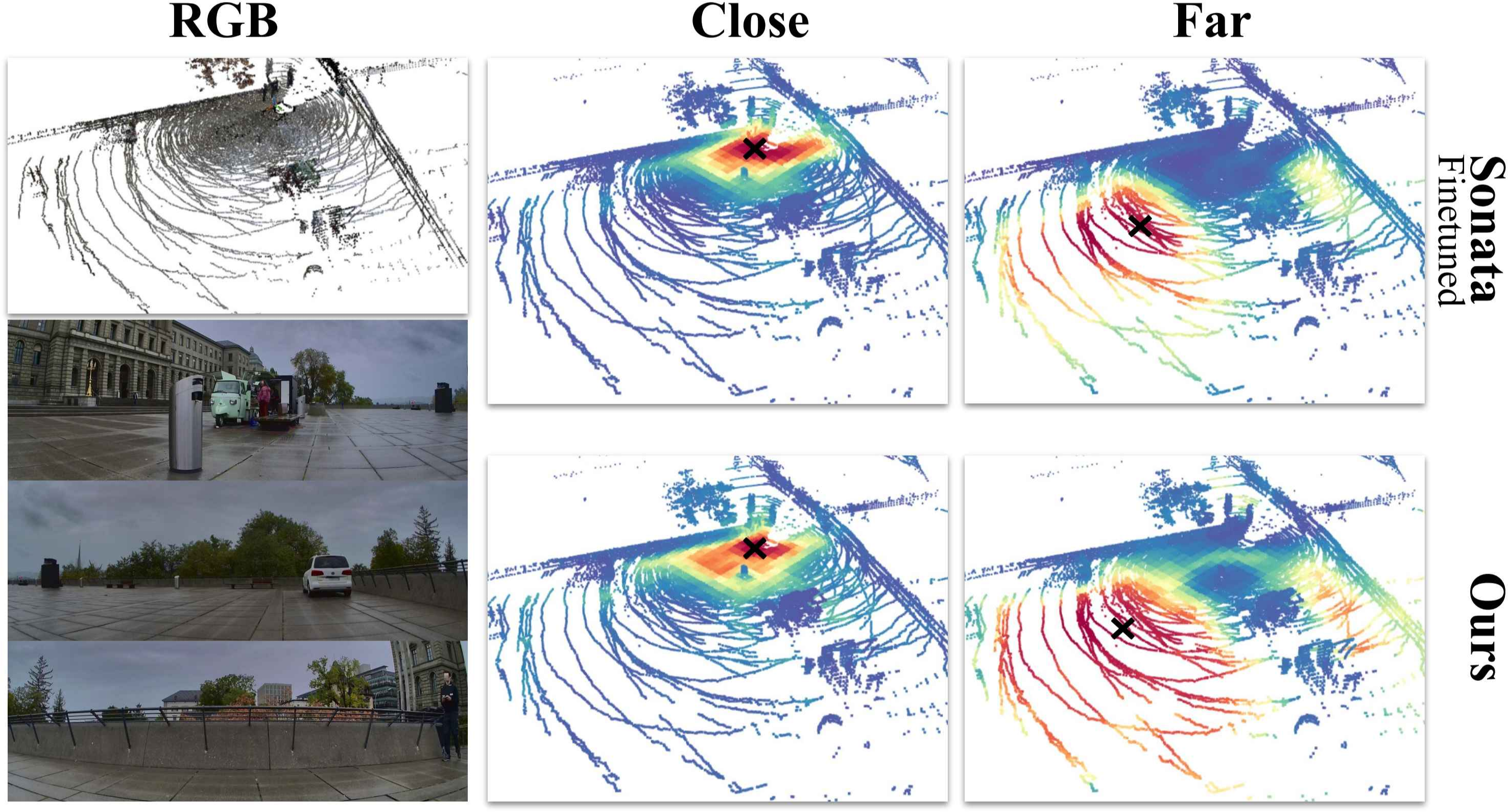}
    \caption{\textbf{Sparsity Encoding Phenomenon}.
    We show cosine similarity maps of a scene (left) for a reference point close (middle) and far away (right) from the robot.
    Despite the uniform terrain at both references, the embeddings diverge.
    Although our method (bottom) demonstrates improved consistency, particularly for distant points, the distinction between dense and sparse regions remains clearly visible.
    Reference highlighted with a cross.
    \vspace{-1em}
    }
    \label{fig:sparsity_encoding}
\end{figure}

Our preliminary exploration indicates that Sonata-style models are highly sensitive to point density variations.
While this issue is negligible for indoor datasets like ScanNet~\cite{dai2017scannet} with uniform density, LiDAR data exhibits a quadratic density decay with range.
The model implicitly encodes local sparsity patterns, introducing a range-dependent bias that causes representations to drift based on their distance from the sensor.
We illustrate this phenomenon in Fig.~\ref{fig:sparsity_encoding}.

To mitigate this, we introduce a Sparse View augmentation.
We simulate sparsity by uniformly subsampling the global view with a ratio $r_s \sim \mathcal{U}(r_{\min}, r_{\max})$.
The student is tasked with aligning its representation of the sparse view with the teacher's dense view.
This sparse-to-dense regularization encourages the model to learn density-invariant features, critical for consistent perception across density ranges.
Formally, we extend the student view set to $\mathcal{V}_S = \mathcal{V}_{\text{mask}} \cup \mathcal{V}_{\text{loc}} \cup \mathcal{V}_{\text{sparse}}$.
Similar to the masked views, sparse views are supervised by the full set of teacher views $\mathcal{A}(s) = \{t_1, t_2\}$ to maximize the supervision signal.
In alignment with Sonata, loss weights are chosen such that each view is weighted equally, and we employ a progressive scheduler, decreasing $(r_{\min}, r_{\max})$ from $(0.9, 1.0)$ to $(0.5, 0.7)$ during training.

\subsection{Resource-Efficient Self-Distillation}
\label{sec:method:self_distillation}
Our objective minimizes the cross-entropy between the student's prototype probability predictions and the teacher's SK-normalized prototype assignments~\cite{caron2020unsupervised}.
While SK normalization prevents mode collapse, it benefits from a large batch size to accurately estimate the global prototype distribution, a scale Sonata achieves using 32 GPUs.
To stabilize training on only 4 GPUs, we utilize a memory bank adapted from SwAV~\cite{caron2020unsupervised}.
We maintain a FIFO queue of prototype scores $S_{\text{bank}}$ and concatenate the current batch's $B$ point-wise scores $S_{\text{batch}}^{(t)}$ with this queue prior to SK normalization:
\begin{equation}
Q^{(t)} = \text{SinkhornKnopp}(\text{Concat}(S_{\text{batch}}^{(t)}, S_{\text{bank}}))[:B].
\end{equation}
This effectively decouples the normalization statistics from the batch size, stabilizing assignment on limited compute.

\subsection{Cross-Modal Distillation}
\label{sec:method:cross_distillation}

To further improve our representations, we add a secondary cross-modal distillation objective.
Leveraging image features derived from frozen Vision Foundation Models offers two advantages: they provide robust representations due to their massive pretraining scales, and they serve as stable targets to guide the 3D student's learning~\cite{puy2024three, zeid2025dino, zhang2026concerto}.
To backproject the image features onto each point, we use known camera intrinsics and extrinsics.
However, perspective projection causes a single low-resolution feature patch to cover an increasingly large 3D volume at long ranges.
Therefore, instead of backprojecting the DINOv2~\cite{oquab2023dinov2} patches directly, we first upsample them with LoftUp~\cite{huang2025loftup} (224~px short side) before bilinearly sampling at the projected point coordinates.
We thus obtain a high-resolution feature $h_j^{(t)}$ per point $j$ and teacher view $t$.
Let $z_i^{(s)}$ be the student's feature associated with point $i$ in student view $s$, projected via a small MLP to match the feature dimension of the teacher, and $\tilde{\cdot}$ denote $\ell_2$-normalization.
Following~\cite{puy2024three}, we define the distillation loss as:

\begin{equation}
\mathcal{L}_{\text{sim}}(s, t) = \frac{1}{\lvert\mathcal{M}_{s,t}\rvert} \sum_{(i,j) \in \mathcal{M}_{s,t}} \lVert \tilde{h}_j^{(t)} - \tilde{z}_i^{(s)}\rVert_2.
\end{equation}
$\mathcal{M}_{s,t} = \{(i,j)\}$ defines the set of nearest-neighbor spatial assignments between the two views.
The total cross-modal loss follows~{(\ref{eq:loss_uni})}:
\begin{equation}
\mathcal{L}_{\text{cross}} = \sum_{s \in \mathcal{V}_S} \lambda_s \sum_{t \in \mathcal{A}(s)} \mathcal{L}_{\text{sim}}(s, t).
\end{equation}

Finally, the overall training objective is a simple sum of both losses, requiring no hyperparameter tuning:
\begin{equation}
\mathcal{L} = \mathcal{L}_{\text{uni}} + \mathcal{L}_{\text{cross}}.
\end{equation}

\section{EXPERIMENTS}
\label{sec:experiments}

\begin{figure*}
    \centering
    \vspace{0.5em}
    \includegraphics[width=0.88\linewidth]{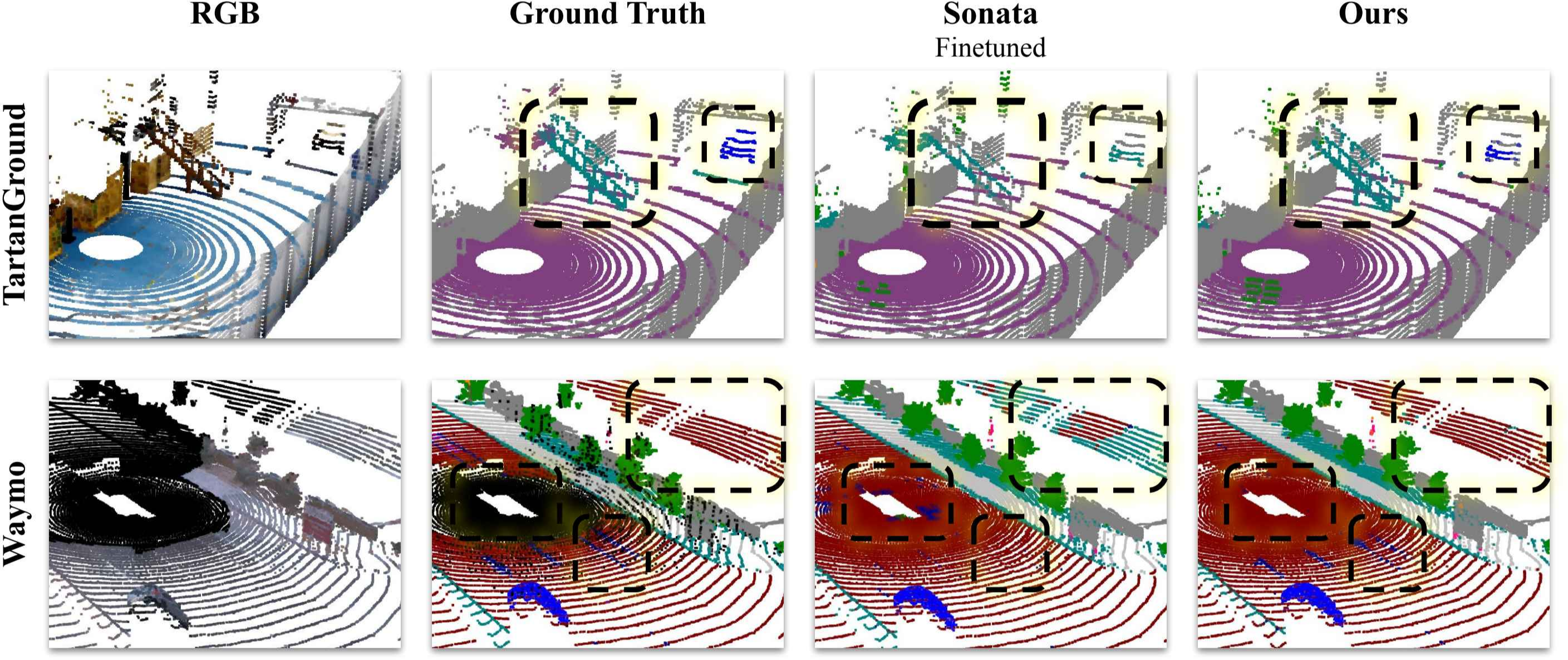}
    \caption{\textbf{Qualitative Results.} 
    We compare qualitative segmentation results against the finetuned Sonata variant on TartanGround (top) and Waymo (bottom).
    In TartanGround, Vernata demonstrates significant improvements in segmenting challenging classes (stairs), alongside better performance in sparse regions (distant vehicle).
    However, both models exhibit limitations in resolving the walkway and occasionally misclassify structural elements as vegetation.
    In Waymo, we achieve more accurate road segmentation especially in the dense and sparse regions, as well as noticeably cleaner lane markers.
    }
    \vspace{-0.2em}
    \label{fig:qualitative}
\end{figure*}

We evaluate our framework across two domains: unstructured field environments and urban driving.
In the former, we combine the real-world GrandTour~\cite{frey2026grandtour} and synthetic TartanGround~\cite{patel2025tartanground} for self-supervised pretraining.
However, as GrandTour lacks semantic annotations, we perform downstream evaluation exclusively on TartanGround.
For the latter, we utilize the Waymo Open Dataset~\cite{sun2020scalability} as a self-contained benchmark, employing it for both pretraining and evaluation to validate our method on an established dataset.

\textbf{TartanGround Standardization.}
As TartanGround lacks a canonical evaluation protocol, we establish one for semantic segmentation.
First, to mitigate LiDAR sparsity, we accumulate point clouds over 3 frames during preprocessing.
Second, we map the 1496 scene-specific raw labels to 7 canonical semantic classes (e.g., \textit{Manmade Surface}, \textit{Natural Surface}, \textit{Static Obstacles}, \textit{Vegetation}) and define a stratified training/validation split.
Third, we establish distinct dataset scales for our two training phases.
To maximize variety during self-supervised pretraining, we utilize a broad dataset of \printval\ttgLengthSSL samples.
For linear probing, we restrict the dataset to a curated subset of \printval\ttgLength samples to ensure the network learns from only the highest-quality semantic labels.

\textbf{Custom Dataset.}
To evaluate the real-world applicability of our model, we introduce a small, in-house annotated dataset. This dataset comprises 311 labeled frames collected across four distinct trail trajectories: two sourced from GrandTour~\cite{frey2026grandtour} (\enquote{Grindelwald - Canyon} 1 and 2) and two that are self-collected.
The labeling consists of 30 classes tailored to our use case, such as \textit{Path}, \textit{Vegetation}, \textit{Person}, or \textit{Skill Element}.
We reserve one of the self-collected trajectories as a hold-out validation set for quantitative evaluation.

\textbf{Implementation Details.} 
We initialize all our models from a ScanNet~\cite{dai2017scannet}-pretrained Sonata checkpoint, as it is the only publicly available pretrained checkpoint at this time.
Subsequently, we first perform self-supervised pretraining on our target datasets, followed by linear probing for semantic segmentation.
We report mean Intersection-over-Union (mIoU), mean Accuracy (mAcc), and overall Accuracy (Acc).
For details regarding hyperparameter setup, please refer to Table~\ref{tab:hyperparameters} in the Appendix.
To manage the significant computational overhead of self-supervised pretraining and validation across diverse datasets, we restrict all point clouds to a 50~m~$\times$~50~m region centered around the sensor, voxelized with a grid size of 0.1~m during training.


\begin{table}[t]
\centering
\caption{\textbf{Semantic Segmentation via Linear Probing.}
We evaluate Vernata against the original Sonata~\cite{wu2025sonata} baseline (pretrained on ScanNet~\cite{dai2017scannet}) and a domain-adapted version of Sonata (self-supervised finetuned on the respective LiDAR datasets).
Performance is reported using mIoU, mAcc, and Acc. By addressing LiDAR sparsity and modality gaps, our approach consistently outperforms both baselines.
}
\label{tab:results_main}
\small 
\setlength{\tabcolsep}{3pt}
\begin{tabular}{lcccccc}
    \toprule
    & \multicolumn{3}{c}{\textbf{TartanGround}} & \multicolumn{3}{c}{\textbf{Waymo}} \\
    \cmidrule(lr){2-4} \cmidrule(lr){5-7}
    Method & mIoU & mAcc & Acc & mIoU & mAcc & Acc \\
    \midrule
    Sonata~\cite{wu2025sonata} & \ttgSonataMIoU & \ttgSonataMAcc & \ttgSonataAcc & \waymoSonataMIoU & \waymoSonataMAcc & \waymoSonataAcc \\
    Sonata~\cite{wu2025sonata} Finetuned & \ttgBaselineMIoU & \ttgBaselineMAcc & \ttgBaselineAcc & \waymoBaselineMIoU & \waymoBaselineMAcc & \waymoBaselineAcc \\
    \textbf{Ours} & \textbf{\ttgUltimateMIoU} & \textbf{\ttgUltimateMAcc} & \textbf{\ttgUltimateAcc} & \textbf{\waymoUltimateMIoU} & \textbf{\waymoUltimateMAcc} & \textbf{\waymoUltimateAcc} \\
    \bottomrule
    
\end{tabular}
\end{table}

\begin{table}[t]
\centering
\caption{\textbf{Ablation of Contributions.}
We ablate the Sparse View (SP), Sinkhorn-Knopp Memory Bank (MB), and Cross-Modal Distillation (CMD) extensions.
Across both datasets, we observe moderate gains using SP and MB, while the largest gains stem from CMD.
Combining all extensions yields the best mIoU across both datasets.
}
\label{tab:ablation}
\small 
\setlength{\tabcolsep}{3pt}
\begin{tabular}{ccccccc}
    \toprule
    SP & MB & CMD & \multicolumn{2}{c}{\textbf{TartanGround}} & \multicolumn{2}{c}{\textbf{Waymo}} \\
    \cmidrule(lr){4-5} \cmidrule(lr){6-7}
     &  &  & mIoU & mAcc & mIoU & mAcc \\
    \midrule
    - & - & - & \ttgBaselineMIoU & \ttgBaselineMAcc & \waymoBaselineMIoU & \waymoBaselineMAcc \\
    \checkmark & - & - & \ttgSparsityMIoU & \ttgSparsityMAcc & \waymoSparsityMIoU & \waymoSparsityMAcc \\
    - & \checkmark & - & \ttgSinkhornKnoppMIoU & \ttgSinkhornKnoppMAcc & \waymoSinkhornKnoppMIoU & \waymoSinkhornKnoppMAcc \\
    \checkmark & \checkmark & - & \ttgSparsitySKMIoU & \ttgSparsitySKMAcc & \waymoSparsitySKMIoU & \waymoSparsitySKMAcc \\
    - & - & \checkmark & \textit{\ttgCMDMIoU} & \textit{\ttgCMDMAcc} & \textit{\waymoCMDMIoU} & \textbf{\waymoCMDMAcc} \\
    \checkmark & \checkmark & \checkmark & \textbf{\ttgUltimateMIoU} & \textbf{\ttgUltimateMAcc} & \textbf{\waymoUltimateMIoU} & \textit{\waymoUltimateMAcc} \\
    \bottomrule
\end{tabular}
\end{table}

\subsection{Main Results}
We benchmark our method against two self-supervised baselines: the official Sonata~\cite{wu2025sonata} checkpoint (frozen, pretrained on ScanNet~\cite{dai2017scannet}) and a Sonata variant finetuned on our respective target datasets.
As shown in Table~\ref{tab:results_main}, our approach consistently outperforms both baselines across all metrics.
On TartanGround, we observe a gain of +\printval\absTtgMIoU mIoU (+\printval\relpercTtgMIoU), while on the Waymo dataset, we improve by +\printval\absWaymoMIoU mIoU (+\printval\relpercWaymoMIoU) over the finetuned baseline.
Qualitative results further reinforce these findings (see Fig.~\ref{fig:teaser} and Fig.~\ref{fig:qualitative}).
Consistent across both datasets, Vernata demonstrates more robust segmentation, especially within the nearby dense and distant sparse regions, while significantly improving the detection of challenging semantic classes.
Note that the quantitative results are not directly comparable to those cited in Sonata~\cite{wu2025sonata}, as we only finetune from a ScanNet~\cite{dai2017scannet} checkpoint instead of training from scratch, consider only a 50~m~$\times$~50~m subsection of the scene, use a coarser downsampling grid for inference, and omit test-time augmentation.

\begin{figure*}[ht]
    \centering
    \vspace{0.5em}
    \includegraphics[width=0.78\linewidth]{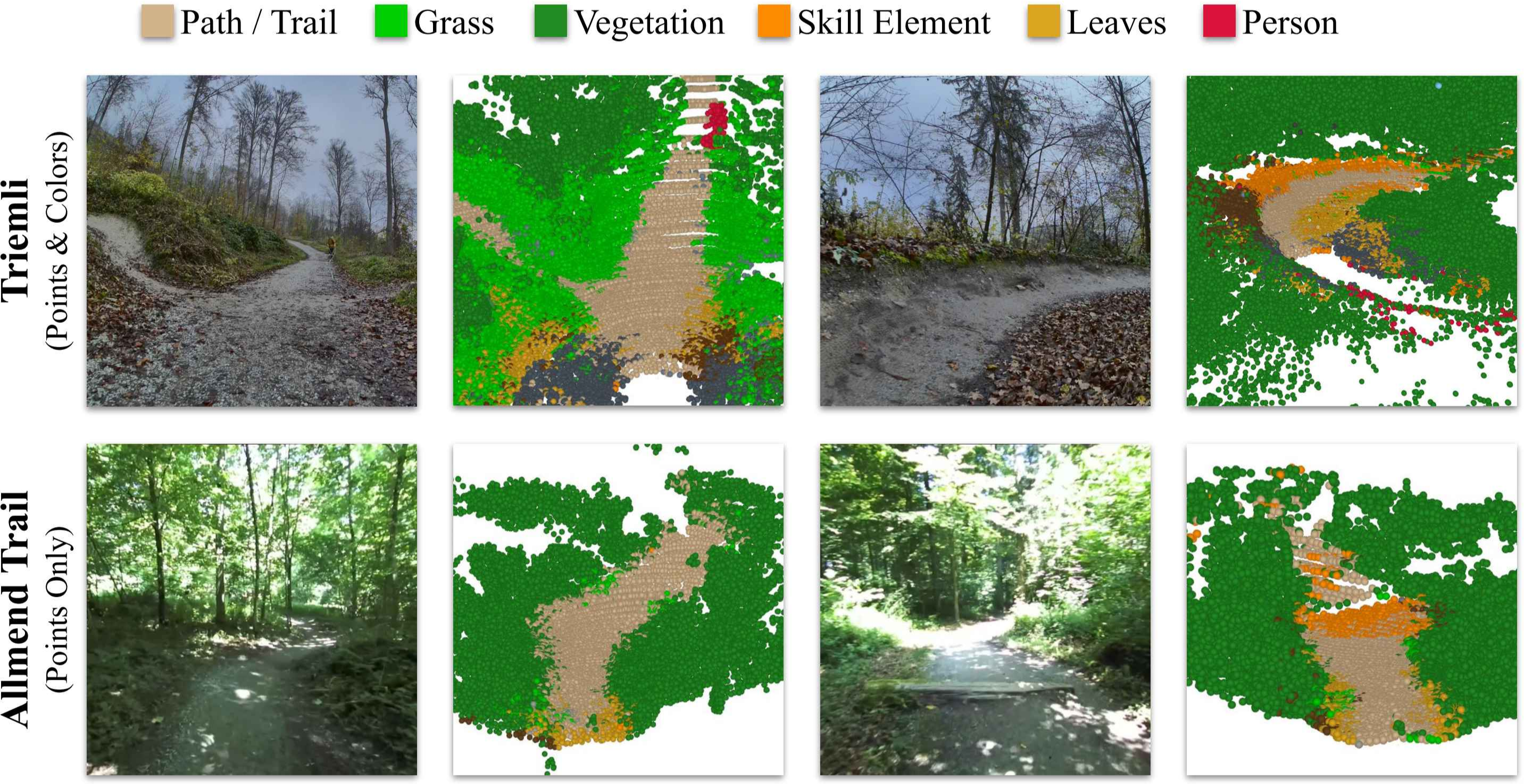}
    \caption{
    \textbf{Inference on Real-World Trajectories.}
    We present inference results for our CMD-less model variant, pretrained on GrandTour and TartanGround, with a linear head finetuned on an in-house dataset of just 311 frames. The top row displays inference on the GrandTour Triemli scene using only coordinates and colors, while the bottom row shows a trajectory from
    RAI's Ultra Mobility Vehicle (UMV)~\cite{rai2026umv}
    using only point coordinates, no colors or surface normals. Visualized point clouds are accumulated over 5 frames. Both models exhibit robust path detection, capturing humans and skill elements despite constrained modalities and lower LiDAR resolution.
    \vspace{-1em}
    }
    \label{fig:custom_dataset_qualitative}
\end{figure*}

\subsection{Ablation on Proposed Extensions}
We investigate the contribution of each component in Table~\ref{tab:ablation}.
The Sparse View (SP) augmentation provides consistent gains, validating its effectiveness in handling the inherent density variations of LiDAR scans.
The Sinkhorn-Knopp Memory Bank (MB) further boosts performance by stabilizing the training signal during small-batch training.
Finally, Cross-Modal Distillation (CMD) yields the single largest improvement across both datasets, confirming that 2D semantic guidance is highly valuable for scene understanding.
It is, however, slightly less effective on TartanGround (+\printval{\absdiff{\ttgCMDMIoU}{\ttgBaselineMIoU}} mIoU) compared to Waymo (+\printval{\absdiff{\waymoCMDMIoU}{\waymoBaselineMIoU}} mIoU), likely due to the domain gap between synthetic images and the real-world DINOv2 pretraining data.
In summary, the final variant incorporating all extensions yields the highest mIoU across both datasets, with only a small mAcc dip on Waymo, which we attribute to a minor class-balancing trade-off where false positives are reduced at the cost of negligible accuracy.

\begin{table}[t]
\centering
\caption{\textbf{Comparison with PTv3 Baseline.}
We compare our self-supervised approach against the fully supervised PTv3 baseline.
$n$ denotes the number of training samples.
While the fully supervised baseline dominates on large-scale datasets like Waymo, our method significantly closes the gap on smaller datasets (TartanGround) and achieves higher mIoU and mAcc in extremely low-data regimes (Custom).
}
\label{tab:ptv3_comparison}
\small 
\setlength{\tabcolsep}{3pt}
\begin{tabular}{lcccccc}
    \toprule
    & \multicolumn{2}{c}{\textbf{TartanGround}} & \multicolumn{2}{c}{\textbf{Waymo}} & \multicolumn{2}{c}{\textbf{Custom}} \\
    & \multicolumn{2}{c}{($n=\ttgLength$)} & \multicolumn{2}{c}{($n=\waymoLength$)} & \multicolumn{2}{c}{($n=\customLength$)} \\
    \cmidrule(lr){2-3} \cmidrule(lr){4-5} \cmidrule(lr){6-7}
    Method & mIoU & mAcc & mIoU & mAcc & mIoU & mAcc \\
    \midrule
    PTv3~\cite{wu2024point} & \textbf{\ttgPTMIoU} & \ttgPTMAcc & \textbf{\waymoPTMIoU} & \textbf{\waymoPTMAcc} & \customPTMIoU & \customPTMAcc \\
    \textbf{Ours} & \ttgUltimateMIoU & \textbf{\ttgUltimateMAcc} & \waymoUltimateMIoU & \waymoUltimateMAcc & \textbf{\customUltimateMIoU} & \textbf{\customUltimateMAcc} \\
    \bottomrule
    
\end{tabular}
\end{table}

\begin{table}[t]
\centering
\caption{\textbf{Input Modality Robustness.}
We evaluate a CMD-less Vernata variant by sequentially ablating surface normals and color features.
Note that to mitigate computational costs, cross-modal distillation was omitted from this set of experiments.
As expected, mIoU consistently decreases as these auxiliary modalities are removed.
However, the models maintain competitive performance overall.
}
\label{tab:modalities}
\small 
\setlength{\tabcolsep}{3pt}
\begin{tabular}{ccccccc}
    \toprule
    Coords & Color & Normals & \multicolumn{2}{c}{\textbf{TartanGround}} & \multicolumn{2}{c}{\textbf{Waymo}} \\
    \cmidrule(lr){4-5} \cmidrule(lr){6-7}
     &  &  & mIoU & mAcc & mIoU & mAcc \\
    \midrule
    \checkmark & \checkmark & \checkmark & \textbf{\ttgSparsitySKMIoU} & \textbf{\ttgSparsitySKMAcc} & \textbf{\waymoSparsitySKMIoU} & \textbf{\waymoSparsitySKMAcc} \\
    \checkmark & \checkmark & - & \ttgCrdClrMIoU & \ttgCrdClrMAcc & \waymoCrdClrMIoU & \waymoCrdClrMAcc \\
    \checkmark & - & - & \ttgCrdMIoU & \ttgCrdMAcc & \waymoCrdMIoU & \waymoCrdMAcc \\
    \bottomrule
\end{tabular}
\end{table}

\subsection{Comparison to Supervised Approaches}
We compare our method to fully supervised approaches to evaluate SSL pretraining in small data regimes.
Table~\ref{tab:ptv3_comparison} compares a fully supervised PTv3~\cite{wu2024point} baseline, trained from scratch, to a linear head trained on our SSL-pretrained models.
Benefiting from a 7.5M-parameter decoder compared to our lightweight 9–37k-parameter linear head, PTv3 heavily outperforms our approach on the larger Waymo dataset ($n=\waymoLength$), leading by \printval{\absdiff{\waymoPTMIoU}{\waymoUltimateMIoU}} mIoU.
%
However, this performance gap narrows significantly on the smaller TartanGround dataset ($n=\ttgLength$). On TartanGround, PTv3's lead shrinks to just \printval{\absdiff{\ttgPTMIoU}{\ttgUltimateMIoU}} mIoU, 
and our SSL approach actually surpasses the fully supervised baseline by \printval{\absdiff{\ttgUltimateMAcc}{\ttgPTMAcc}} mAcc.
Finally, on our custom dataset with very few examples ($n=\customLength$), the SSL approach demonstrates its strength in low-data regimes by outperforming the fully supervised baseline by \printval{\absdiff{\customUltimateMIoU}{\customPTMIoU}} mIoU and \printval{\absdiff{\customUltimateMAcc}{\customPTMAcc}} mAcc.
These results demonstrate that as the availability of annotated data decreases, the representations learned during large-scale pretraining allow SSL models to surpass supervised baselines. 
To further illustrate this real-world generalization, Fig.~\ref{fig:custom_dataset_qualitative} presents real-world inference results from our CMD-less model finetuned on this custom dataset. 
Whether evaluated on a GrandTour trail (top row) or a trajectory captured via
RAI's UMV~\cite{rai2026umv}
(bottom row), the model exhibits robust path detection and accurately captures dynamic obstacles like humans, even when operating under constrained modalities without normals or color.

\subsection{Robustness to Reduced Modalities}
The standard architecture expects 3 inputs: coordinates, colors, and surface normals.
However, in robotics, high-quality normals or omnidirectional calibrated camera coverage is often unavailable.
Thus, we evaluate our model trained on a reduced modality set. Because we omit cross-modal distillation for this experiment due to the additional training overhead, we refer to this as the \mbox{CMD-less} variant, which is also showcased in Fig.~\ref{fig:custom_dataset_qualitative}.
Encouragingly, Table~\ref{tab:modalities} shows that our model retains strong performance even without colors or normals, achieving \printval\ttgCrdMIoU~mIoU on TartanGround and \printval\waymoCrdMIoU~mIoU on Waymo using only coordinates.

\subsection{Comparison to Patch-based Distillation}
We compare the patch-based distillation mechanism proposed in Concerto~\cite{zhang2026concerto} to our high-resolution feature distillation.
As shown in Table~\ref{tab:distillation_mechanism}, both methods perform on par overall, with marginal trade-offs across datasets.
However, it also reveals a performance shift at longer ranges: when evaluating exclusively on LiDAR points beyond a 20-meter radius, our mechanism demonstrates an advantage.
We attribute this to our high-resolution matching mitigating the geometric footprint expansion of low-resolution patches at a distance.
This demonstrates that upsampled feature association holds promise for improving point representations especially at long ranges, and we leave a comprehensive investigation into these architectural trade-offs for future work.

\begin{table}[t]
\centering
\vspace{0.7em}
\caption{
\textbf{Comparison of Distillation Mechanism.}
To compare our high-resolution feature distillation against Concerto's~\cite{zhang2026concerto} patch-based approach, we evaluate models pretrained with only the Cross-Modal Distillation extension.
While both approaches perform on par overall, our dense matching is more robust at longer ranges, achieving higher mIoU on LiDAR points beyond 20 meters (denoted as $20$m+) in both TartanGround (TG) and Waymo.
}
\label{tab:distillation_mechanism}
\footnotesize 
\setlength{\tabcolsep}{3pt}
\begin{tabular}{lcccc|cccc}
    \toprule
     & \multicolumn{2}{c}{\textbf{TG}} & \multicolumn{2}{c}{\textbf{Waymo}} & \multicolumn{2}{c}{$\textbf{TG}_{20\text{m+}}$} & \multicolumn{2}{c}{$\textbf{Waymo}_{20\text{m+}}$} \\
    \cmidrule(lr){2-3} \cmidrule(lr){4-5} \cmidrule(lr){6-7} \cmidrule(lr){8-9}
    Method & mIoU & mAcc & mIoU & mAcc & mIoU & mAcc & mIoU & mAcc \\
    \midrule
    Patch & \textbf{\rangettgConcertoMIoU} & \rangettgConcertoMAcc & \textbf{\rangewaymoConcertoMIoU} & \textbf{\rangewaymoConcertoMAcc} & \rangettgConcertoDistMIoUAboveTwenty & \textbf{\rangettgConcertoDistAccAboveTwenty} & \rangewaymoConcertoDistMIoUAboveTwenty & \rangewaymoConcertoDistAccAboveTwenty \\
    \textbf{Ours} & \textbf{\rangettgMineMIoU} & \textbf{\rangettgMineMAcc} & \rangewaymoMineMIoU & \rangewaymoMineMAcc & \textbf{\rangettgMineDistMIoUAboveTwenty} & \rangettgMineDistAccAboveTwenty & \textbf{\rangewaymoMineDistMIoUAboveTwenty} & \textbf{\rangewaymoMineDistAccAboveTwenty} \\
    \bottomrule
\end{tabular}
\vspace{-0.7em}
\end{table}

\section{CONCLUSION}
In summary, this work introduces Vernata, a framework for self-supervised learning on outdoor LiDAR point clouds.
By extending the Sonata architecture with sparse view augmentation, Sinkhorn-Knopp memory banks, and cross-modal distillation, we effectively address the unique challenges of the outdoor domain.
Our evaluations show that combining these techniques maximizes performance across diverse datasets, with distillation proving exceptionally performant on real-world data like the Waymo Open Dataset.
Furthermore, ablation studies confirm that Vernata maintains robust inference even in limited-modality settings where color and surface normals are unavailable.
While our approach demonstrates clear improvements, it currently operates strictly on a per-frame basis. Future work will focus on integrating temporal context to ensure inference consistency and deploying these robust representations for execution on mobile robots.

\balance
\addtolength{\textheight}{-0cm}

\section*{APPENDIX}

\textbf{Inference.}
On an NVIDIA A5000, we achieve inference rates (FP32 / FP16) of 8.9 / 16.1~Hz on Waymo, 7.2 / 13.3~Hz on TartanGround, and 8.8 / 16.8~Hz on our own dataset.
\vspace{1em}

\textbf{Experiment Setup.} Table~\ref{tab:hyperparameters} shows our hyperparameters.

\begin{table}[h!]
    \centering
    \vspace{0.6em}
    \caption{
        \textbf{Hyperparameter Setup.}
    }
    \begin{tabular}{|c|c|c|}
        \toprule
            \textbf{Training Stage} & \textbf{Parameter} & \textbf{Value}  \\
        \midrule
            \multirow{13}{*}{Self-Supervised}
            & Total Steps & 20,000 \\
            & \multirow{2}{*}{Batch Size} & 24 (Per GPU: 1,\\
            & & \#GPU: 4, GradAcc: 6) \\
            & GPU & NVIDIA H100 80GB \\
            & Learning Rate & 0.0002 \\
            & Schedule & Cosine with warmup \\
            & Global View Size & (0.7, 1.0) $\rightarrow$ (0.4, 1.0) \\
            & Local View Size & (0.1, 0.4) \\
            & \multirow{2}{*}{Masked View} & Size: 0.1 $\rightarrow$ 0.4 \\
            &  & Ratio: 0.3 $\rightarrow$ 0.7 \\
            & Sparse View & (0.9, 1.0) $\rightarrow$ (0.5, 0.7) \\
            & \multirow{2}{*}{Memory Bank} & 50,000 (GT, TG) \\
            & & 100,000 (Waymo) \\
        \midrule
            \multirow{6}{*}{Linear Probing}
            & \multirow{2}{*}{Training Steps} & 10,000 (TG, Custom) \\
            & & 20,000 (Waymo) \\
            & Batch Size & 24 \\
            & Learning Rate & 0.001 \\
            & Schedule & Cosine with warmup \\
            & Loss & CE \& Lovasz~\cite{berman2018lovasz} \\
        \bottomrule
            
    \end{tabular}
    \label{tab:hyperparameters}
\end{table}

\bibliographystyle{IEEEtran}
\bibliography{refs}

\end{document}